%% file: root.tex
\documentclass[letterpaper, 10 pt, conference]{ieeeconf}  % Comment this line out if you need a4paper

\IEEEoverridecommandlockouts                              % This command is only needed if 
\usepackage{subfigure}
\usepackage{amsmath} % assumes amsmath package installed
\usepackage{amssymb}  % assumes amsmath package installed
\usepackage{stmaryrd}   
\usepackage{bm}
\usepackage{url}
\usepackage{graphicx}
\usepackage{float}
\usepackage{siunitx}
\usepackage{threeparttable}
\usepackage{cleveref}
\usepackage{multicol}
\usepackage{booktabs}
\usepackage{multirow}
\usepackage{endnotes}
\usepackage{xcolor}
\usepackage{amssymb}

\newcommand{\sysname}{\texttt{DC-Loc}}

\crefname{section}{\S}{\S\S}
\Crefname{section}{\S}{\S\S}

\newcommand{\ie}{{\em i.e.}}
\newcommand{\eg}{{\em e.g.}}
\newcommand{\et}{{\em et al.}}

\def\spth{\textsuperscript{th}}

\begin{document}
	
\title{\LARGE \bf
	DC-Loc: Accurate Automotive Radar Based Metric Localization with \\Explicit Doppler Compensation \vspace{-15pt}
}	
	
\author{
	Pengen Gao{$^{^\dag}$}, Shengkai Zhang{$^{^\ddag}$}, Wei Wang{${^\ast}^{^\dag}$}, Chris Xiaoxuan Lu{${^\S}$} \vspace{1pt}\\ 
	{$^{^\dag}$}Huazhong University of Science and Technology, {$^{^\ddag}$}Wuhan University of Technology, {${^\S}$}University of Edinburgh 
	\thanks{${^\ast}$The corresponding author is Wei Wang.}%
	\vspace{-10pt}
} % <-this % stops a space

\maketitle
\thispagestyle{empty}
\pagestyle{empty}

%%%%%%%%%%%%%%%%%%%%%%%%%%%%%%%%%%%%%%%%%%%%%%%%%%%%%%%%%%%%%%%%%%%%%%%%%%%%%%%%

\input{sections/abstract}
\input{sections/introduction}

\input{sections/related_work}

\input{sections/methods}
\input{sections/experimental_setup}

\bibliography{ref}
\bibliographystyle{IEEEtran}

\end{document}

%% file: sections/abstract.tex
\begin{abstract}
Automotive mmWave radar has been widely used in the automotive industry due to its small size, low cost, and complementary advantages to optical sensors (\eg, cameras, LiDAR, \textit{etc.}) in adverse weathers, \eg, fog, raining, and snowing. On the other side, its large wavelength also poses fundamental challenges to perceive the environment. Recent advances have made breakthroughs on its inherent drawbacks, \ie, the multipath reflection and the sparsity of mmWave radar's point clouds. However, the frequency-modulated continuous wave modulation of radar signals makes it more sensitive to vehicles' mobility than optical sensors. This work focuses on the problem of frequency shift, \ie, the Doppler effect distorts the radar ranging measurements and its knock-on effect on metric localization. We propose a new radar-based metric localization framework, termed \sysname, which can obtain more accurate location estimation by restoring the Doppler distortion. Specifically, we first design a new algorithm that explicitly compensates the Doppler distortion of radar scans and then model the measurement uncertainty of the Doppler-compensated point cloud to further optimize the metric localization. Extensive experiments using the public nuScenes dataset and CARLA simulator demonstrate that our method outperforms the state-of-the-art approach by $25.2\%$ and $5.6\%$ improvements in terms of translation and rotation errors, respectively.
\end{abstract}

%% file: sections/introduction.tex
\section{Introduction}
%% importance of the studied task
The key to realizing mobile autonomy lies in reliable metric localization. A reliable metric localization system finds a multitude of applications ranging from vehicle navigation to intelligent driving assistance \cite{bresson2017simultaneous}. While the existing approaches based on cameras and LiDARs become increasingly established in ideal conditions, their localization performance falls quickly under visual degradation in challenging environments, such as bad weather, solar glare, and dust. 

%% conceptual opportunity
As an alternative sensor to cameras or LiDARs, millimeter (mmWave) radar is impervious to a variety of visual degradation due to its larger wavelength and stronger ability of signal penetration. Moreover, today's radars often have a longer sensing range compared with optical sensors, lending themselves a robust sensor attractive to many automobile manufacturers. 

\begin{figure}[!t]
	\includegraphics[width=\linewidth]{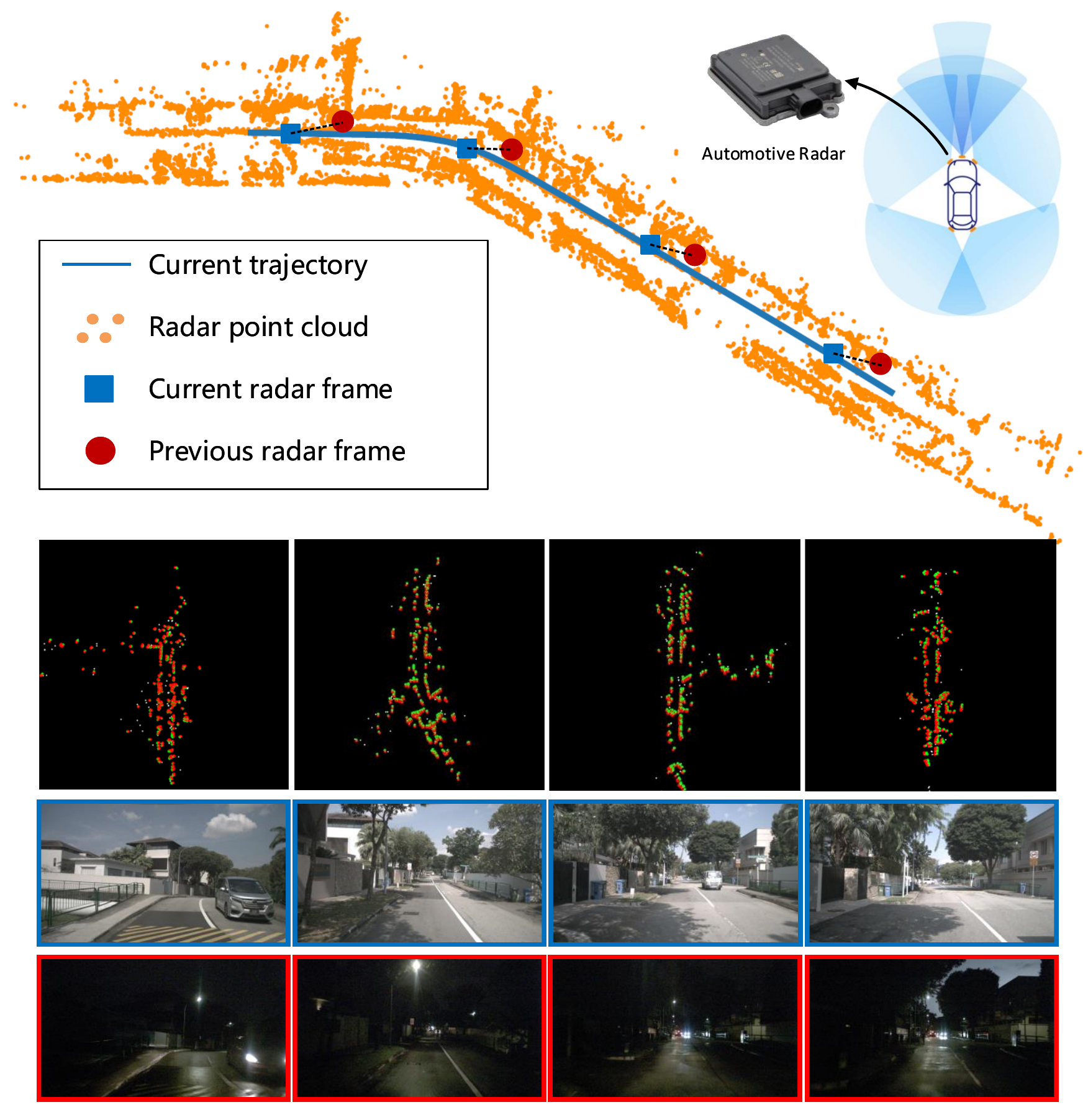}
	\caption{Metric localization using automotive radar: when a vehicle meets a loop closure, the transformation between the current frame (highlighted in \textcolor[rgb]{0,0,1}{blue}) and the previous (highlighted in \textcolor[rgb]{1,0,0}{red}) can be derived from radar submap registration (top row). Such a process can effectively reduce accumulation errors. The middle and bottom rows depict the loop closure images taken by cameras in the daytime and night.}
	\label{fig:openfig}
	\vspace{-20pt}
\end{figure}

%% problems of prior art
Driven by the promise of radar, recent efforts \cite{barnes2020under, tang2020self, wang2021radarloc, de2020kradar++} attempted to use mmWave radars to address the metric localization problem for automobiles. However, these methods do not take the Doppler shift into account and undertreat the resultant ranging error when a vehicle moves at different speeds. As the ranging measurements from most radars are based on the Frequency-Modulated Continuous Wave (FMCW) technique (c.f. Sec.~\ref{sub:radar_principle}), even a slight frequency shift of a wave caused by the Doppler effect will result in a non-negligible drift of ranging measurements \cite{rao2017introduction}. Inconsistent radar scans will be inevitably incurred if such drifts are distinct due to different ego-velocities of two scanning instants. Unfortunately, as the data association step in most metric localization methods relies on the scan consistency, the Doppler effect can substantially threaten the localization success in the case of large ego-velocity difference, e.g., on the highway road. Although  Burnett~\et~\cite{burnett2021we} discussed the influence of Doppler distortion and modeled it into their algorithm, the Doppler effect still has remnant impacts on the localization performance because the spinning radar considered in \cite{burnett2021we} cannot provide accurate Doppler measurement but has to \emph{indirectly} estimate it from other sources.
Empirically we found that the Doppler error still grows larger with the increase of velocity difference between radar scans, even though the method proposed in \cite{burnett2021we} is adopted.

In order to entirely eliminate the Doppler distortion, in this work, we propose to use the emerging automotive mmWave radar (aka. single-chip beamforming radars) for accurate metric localization. For readability, we refer to such radars as automotive radars hereafter. 

Compared with mechanical radars, automotive radars are cheaper and smaller, making them more suitable for autonomous vehicles \cite{kung2021normal}. Besides, the automotive radars can \emph{explicitly} return a richer set of measurements, including the velocity and radar cross-section values of ambient objects (c.f. Fig.~\ref{fig:openfig}). As automotive radar uses beamforming rather than mechanical spinning to scan the environment, it also has a much higher update rate per scan and thus being insusceptible to motion distortion \cite{burnett2021we}.

However, along with the advantages of automotive radars come with low-quality scans. It has been found that this type of radars suffers from limited-angle resolution, higher noise floor, and sparser point cloud density than the mechanical spinning radar \cite{kung2021normal}. The low-quality radar scans, however, threaten the feature association between two scans and undermine the reliability of metric localization when using automotive radars. 

To address the above challenges, we first propose a new method, \sysname to compensate for the Doppler effect on radar ranging measurements by using the off-the-shelf velocity returns from automotive radars. 
Furthermore, we also propose an uncertainty-aware method to mitigate the impact of low-quality radar scans on end localization. Extensive experimental results on the real-world nuScenes dataset and the synthetic CARLA dataset demonstrate the significant performance gain brought by our system, especially when the ego-velocity difference between two scans is large.

%% file: sections/related_work.tex
\section{Related Work}
%Over the years, automotive radar have been widely deployed on vehicles. 
%The industry usually takes RF radars as secondary sensors that assist cameras or Lidars for emergency braking. However, the large wavelength of RF signals holds the capability of combating adverse weather conditions such as foggy, raining, and snowing. Recent advances of Frequency Modulated Continuous Wave (FMCW) radar technology aim to develop such capabilities and they greatly improve the sensing accuracy and range of mmWave radar.

The mainstream research relies on optical sensors such as cameras and LiDARs for metric localization. However, their performance heavily degenerates under challenging environments, \eg, foggy or raining weather. On the contrary, the sensing capability of mmWave radar is immune to visual limitations thanks to its larger wavelength. Moreover, the recent advances in the FMCW radar technology significantly improve the sensing accuracy and range, making the radar a promising sensor for metric localization.

%However, with the advances of Frequency Modulated Continuous Wave (FMCW) radar technology, the sensing accuracy and range of MWR have been greatly improved. These technology advances aim to explore the RF sensing ability in combating adverse weather conditions such as foggy, raining, and snowing. 

%Odometry and localization have recently been the central focus of radar-based navigation research. This work mainly focuses on the improvement of radar metric localization, but since it has similar technical details to Radar Odometry (RO), so we will talk about them together in related work.

\subsection{Optical sensor based Metric Localization}
Camera is a primary sensor for vehicles’ metric localization due to its low cost and general use. Accurate localization results can be obtained by associating camera images with already built maps~\cite{campos2021orb} or existing street views~\cite{agarwal2015metric}. However, such methods are sensitive to illumination changes and textureless environments.

LiDAR, which emits modulated lasers for perception, also attracts much attention in academia and industry. Existing LiDAR systems rely on geometrical constraints of features in point cloud~\cite{shan2020lio, shan2018lego} for motion estimation and metric localization. Meanwhile, learning-based methods have also been well studied and achieve decent performance in ideal conditions~\cite{lu2019l3, Lu_2019_ICCV}. These systems, however, are vulnerable to dust or smoke. They perform poorly in adverse weather conditions.

%combine the radar and Lidar for localization, which
\subsection{Radar-based Metric Localization}

RF signals can penetrate, reflect, or diffract from obstacles, making mmWave radars robust to visual degradation. Meanwhile, the signal’s large wavelength poses many fundamental challenges of mapping and localization, \eg, multipath reflections and low-resolution perceptions. Prior arts have made tremendous efforts to tackle such problems.

Ward~\et~\cite{ward2016vehicle} propose a two-stage pipeline for the metric localization using radars. They adopt an EKF-based framework that integrates the Iterative Closest Point (ICP)~\cite{besl1992method} algorithm with an existing map. RadarSLAM~\cite{hong2020radarslam} demonstrates the first radar-based Simultaneous Localization and Mapping (SLAM) system that works well in adverse weathers, utilizing vision-based schemes for feature extraction and graph matching. Barnes~\et~\cite{barnes2020under} present a correlation-based framework to detect key points and generate global descriptors from radar scans, significantly improving the odometry estimation accuracy. 
% \pe{Wei~\et~\cite{wang2021radarloc} devise a geometry-aware neural network, which estimates the global 6-DoF pose from a single radar scan by leveraging self-attention.}
Park~\et~\cite{park20213d} use a customized hardware for data collection and model the radar SLAM problem into a pose-graph framework, which substantially improves the performance over the single sensor case in the existence of dynamic objects.
% Daniele~\et~\cite{de2020kradar++} propose a coarse-to-fine radar localization algorithm that consolidates the place recognition and radar pose estimation algorithms into a hierarchical localization process.}

More efforts explore cross-modality solutions.
Yin~\et~\cite{yin2020radar} combine the radar and LiDAR outputs, transferring raw radar scans to synthetic LiDAR images and adopting the Monte Carlo algorithm to localize the vehicle. Tang~\et~\cite{tang2020self} propose an unsupervised-learning framework that aligns 2D radar scans with the geometric structure in satellite images. Although prior works have made performance improvements in metric localization, they neglect the impact of the Doppler effect that severely undermines the metric localization, especially at high speeds.

The closest work to ours is~\cite{burnett2021we}. It mitigates the Doppler distortion in the metric localization using spinning radars. Their Doppler compensation scheme has been the state-of-the-art (SOTA) solution for radar-based metric localization. However, due to the limitation of spinning radar, the system infers the velocity from the pose estimates, which have been corrupted by the Doppler shift. In practice, we find that their metric localization error still grows larger with the increase of velocities between radar scans. 

Our work focuses on the problem of explicitly eliminating the Doppler effect in the metric localization using automotive radars. Combining with an uncertainty-aware optimization scheme,  our system can better handle the high noise floor of the measurements from low-cost automotive radars.

%% file: sections/methods.tex
\section{Preliminary}
\begin{figure}[h]
	\includegraphics[width=\linewidth]{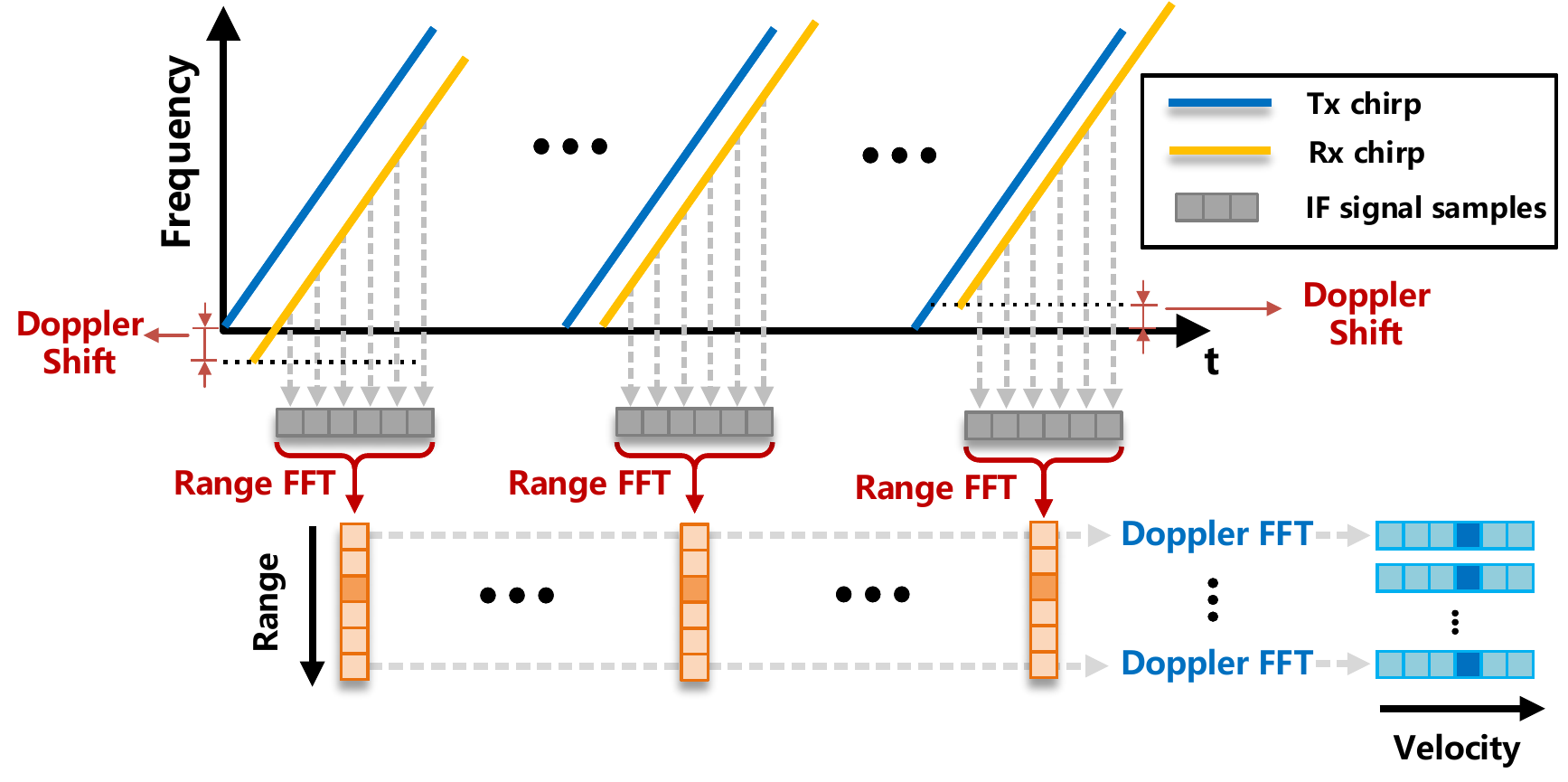}
	\caption{The process of FMCW-based range and velocity measurements. The Doppler frequency shift occurs when the radar and objects in the environment have relative velocities. The frequency shift transforms into a range offset to radar measurements.\vspace{-10pt}}
	\label{fig:mmWave_principle}
\end{figure}

\subsection{FMCW Radar Principle}
\label{sub:radar_principle}
Automotive mmWave radars usually adopt FMCW signals for measuring the range (\ie, the point cloud in a scan) and relative radial velocity between the vehicle and a target object. When the radar receives a reflected signal from a target, it performs a dechirp operation by mixing the received signal with the transmitted signal, which produces an \textit{Intermediate Frequency} (IF) signal $s(t)$, \ie,
\begin{equation}
	s(t) = A\exp[j(2\pi f_bt + \psi_l)], \label{eq:1} 
\end{equation}
where $A$ denotes the signal attenuation, $f_b$ and $\psi_l$ are the frequency and the phase corresponding to the central frequency of the IF signal, respectively. The distance and radial velocity between the radar and its target can be calculated as~\cite{ding2016vibration}:
\begin{equation}
	R_m = \frac{f_b c}{2K},\ \   
	v_m = \frac{f_v\lambda}{2 T_c} \ \ (K=B/T_r),
	\label{eq:2} 
\end{equation}
where $B$, $T_r$, and $\lambda$ denotes bandwidth, signal duration, and signal wavelength. $f_v$ and $T_c$ denote phase change rate and the time separation between IF signals. 

\subsection{Doppler Effect in Range Measurement}
The above ranging operation of FMCW radars works well when a vehicle's ego-velocity is low. However, for autonomous driving scenarios where the ego-velocity changes dramatically from time to time, the Doppler shift on the spectrum will incur a non-negligible knock-on effect on the range measurements. As shown in Fig.~\ref{fig:mmWave_principle}, the Doppler effect causes an apparent frequency shift on the FMCW signal, which eventually results in the Doppler distortion to the point cloud from a radar scan~\cite{burnett2021we}. 
% Such a distortion is very likely to mismatch a loop closure, significantly degrading the performance of metric localization.
Such a distortion significantly degrades the performance of metric localization.

\section{Methods}
This section first models the Doppler shift in radar scans and elaborates on the issue of a vehicle's metric localization. To address the issue, we present two designs in \sysname: 1) the Doppler compensation approach to restore the Doppler distortion; 2) the uncertainty estimation scheme to optimize the metric localization.

\subsection{Doppler Shift Modeling}
%\noindent\textit{1) Doppler offset model} 
The Doppler shift $f_d = \frac{2v_r}{\lambda}$, where $v_r$ denotes the radial component of the relative velocity and $\lambda$ denotes the signal wavelength. Thus, the IF signal frequency $f_b$ is distorted as $\hat{f}_b = f_b + f_d$, producing a range shift $r_d$ to the range $r$. The measured range can be expressed as 
\begin{equation}
	\hat{r} = r + r_d, \ 
	r_d = \frac{f_c}{K}v_r, \; K=B/T_r,
	\label{eq:3}
\end{equation}
where $f_c$ is the central frequency, $B$ is the bandwidth, and $T_r$ is the signal duration. Thus, $K$ represents the ramp rate of a chirp. In addition, the range shift $r_d$ is proportional to the radial velocity $v_r$, indicating that approaching a target in environments shortens the range and vice versa. Although $r_d$ is also inversely proportional to $K$, mmWave radars typically keep a low ramp rate to guarantee its range resolution (small $K$)~\cite{rao2017introduction}, making velocity $v_r$ the dominant factor distorts a radar's range scan. 

To our knowledge, there is no elegant solution to address the Doppler distortion for single-chip mmWave radars. The closest work~\cite{burnett2021we} is designed for spinning mmWave radars. Since spinning radars cannot measure the radial velocity due to its progressive data collection mechanism, the velocity is derived from the vehicle's ego-motion, whose accuracy has been affected by the Doppler distortion.

\begin{figure}
	\includegraphics[width=\linewidth]{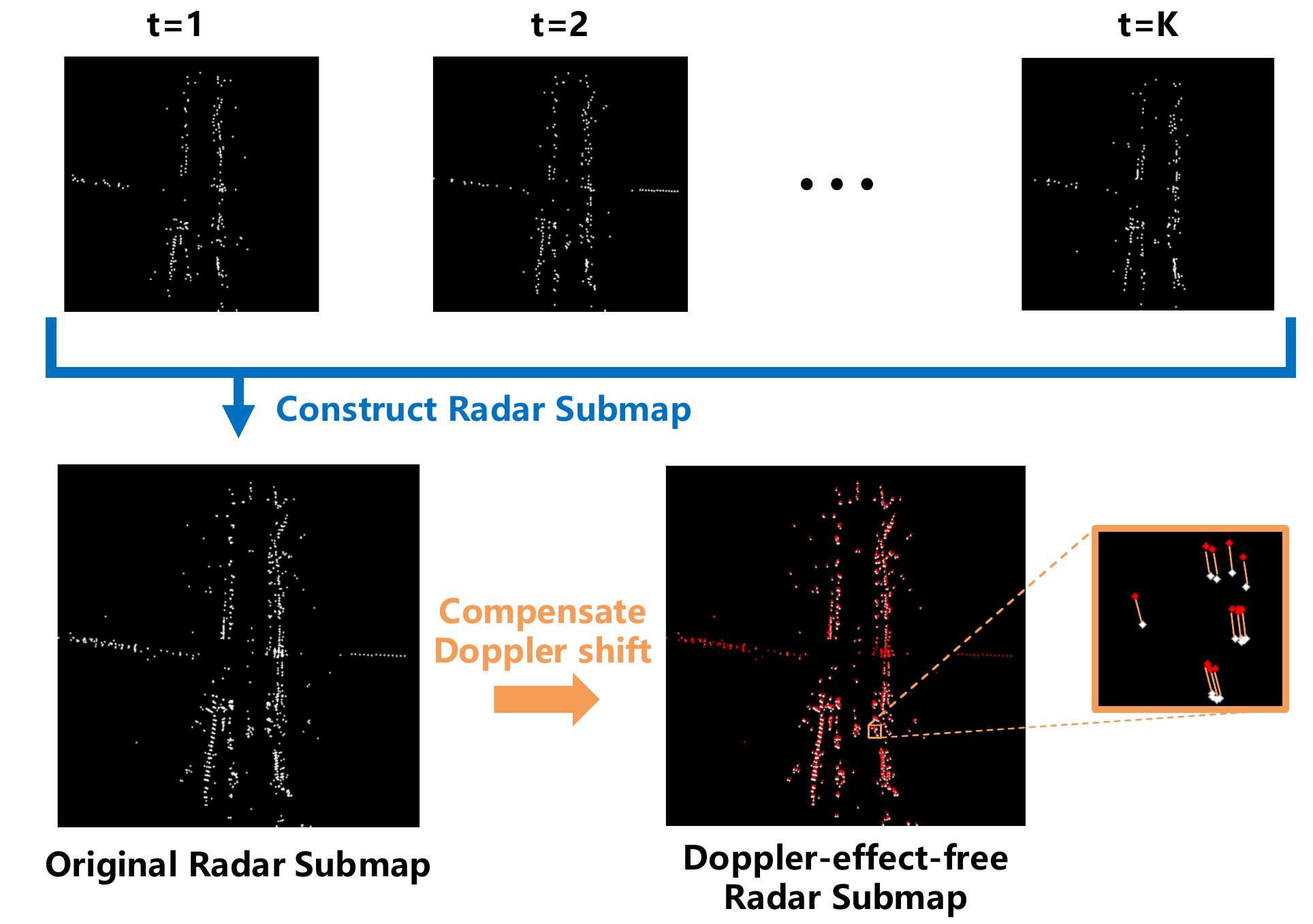}
	\caption{An example of our radar submap (bottom left) built from $K$ radar scans (top images): The Doppler compensation is performed in each scan to restore the radar submap (bottom right), where the points after compensation are highlighted in \textcolor[rgb]{1,0,0}{red}. The \textcolor[rgb]{1, 0.5, 0}{orange} lines represent for $r_d$. \vspace{-15pt}}
	\label{fig:submaps}
	
\end{figure}

\subsection{Doppler Shift Compensation} 

The key idea behind our Doppler compensation method is that the radial velocity measured by automotive mmWave radars is resilient to the Doppler frequency shift. Recent work~\cite{skolnik2008radar} concludes that the Doppler effect can only shift the frequency peaks in the range bin rather than their phases. And the phases are essential to estimate the radial velocity. Specifically, as shown in Fig.~\ref{fig:mmWave_principle}, the Doppler FFT utilizes the phases of the range bins to estimate the velocity. The FFT outputs provide the phase change rate of the range bins, having no relationship with the frequency of the IF signals. Thus, the Doppler effect has no impact on the velocity measurements.

%The FFT outputs provide the phase change rate of the range bins, which has no relationship with their frequency (Eqn.~\eqref{eq:1}), meaning that the Doppler effect has no impact.

%\pe{To eliminate the Doppler distortion in the radar scans, }
Based on the above observation, we now elaborate on the Doppler shift compensation. Suppose there are $M$ mmWave radars installed on the vehicle to perceive environments. We denote the point cloud and radial velocity measurements of $m$\spth radar in Cartesian coordinate as $\hat{\mathcal{P}}_m = \{\hat{\bm{p}}_{mi}|i=1,2,\cdots,N_m\},\; \hat{\mathcal{V}}_m=\{\hat{\bm{v}}_{mi}|i=1,2,\cdots,N_m\}$, $m = 1, 2, \cdots, M$, where $N_m$ is the number of targets in the radar scan. Utilizing the velocities can restore the targets' positions from $m$\spth radar as $\bar{\mathcal{P}}_m$:
\begin{equation}
	\begin{aligned}
		\bar{\mathcal{P}}_m &= \{\bar{\bm{p}}_{mi}|i=1,2,...N\}, \\
		\bar{\bm{p}}_{mi} &= \hat{\bm{p}}_{mi}-\frac{f_c}{K} \hat{\bm{v}}_{mi}, \  \hat{\bm{p}}_{mi} \in \hat{\mathcal{P}}_m, \ \hat{\bm{v}}_{mi} \in \hat{\mathcal{V}}_m. 
	\end{aligned}
	\label{eq:4}
\end{equation}
Then we can obtain a full Doppler-compensated radar scan $\bar{\mathcal{P}}$ by transforming each $\bar{\mathcal{P}}_m$ into the inertial frame:
\begin{equation}
	\begin{aligned}
		\bar{\mathcal{P}}_m &= \{\bm{T}_{m}^{b}\bar{\bm{p}}_{m,i}|i=1,2,...N\}, \ m = 1,2,...M, \\
		\bar{\mathcal{P}} &= \bar{\mathcal{P}}_1 \cup \bar{\mathcal{P}}_2 \cup ... \cup \bar{\mathcal{P}}_M, 
	\end{aligned}
	\label{eq:5}
\end{equation}
where $\bm{T}_{m}^{b} \in SE(3)$ represents the extrinsic parameters of $m$\spth radar. 

However, a single radar scan can hardly provide enough information for metric localization. As illustrated in Fig.~\ref{fig:submaps}, in order to overcome the sparsity of the mmWave radar's outputs, we concatenate several consecutive Doppler-compensated radar scans together to get a denser radar submap. We denote such superposition point cloud as a Doppler-effect-free radar submap, which is more informative and accurate for the subsequent steps.

\begin{figure}[t]
	\includegraphics[width=\linewidth]{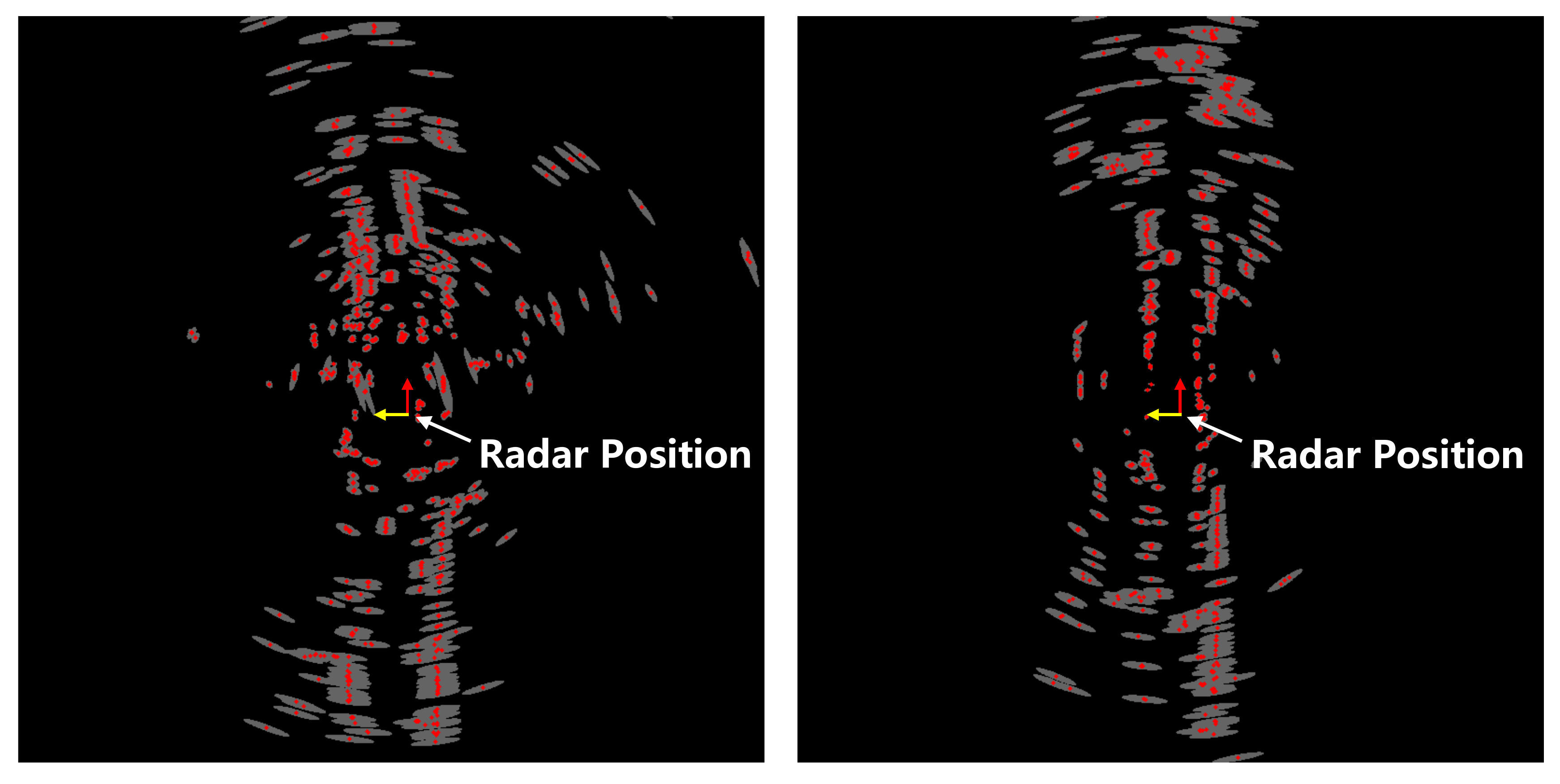}
	\caption{The uncertainty estimation for each point (\textcolor[rgb]{1,0,0}{red}) with their 5 standard deviation uncertainty ellipse (\textcolor[rgb]{0.4,0.4,0.4}{gray}). The points further away from the radar (image center) have larger uncertainty ellipses.}
	\label{fig:uncertainty}
	\vspace{-10pt}
\end{figure}

\subsection{Uncertainty-aware Metric Localization}
Although the Doppler distortion has been restored in the above compensation method, the measured velocities and ranges still have significant uncertainties due to the low-quality radar outputs, especially when targets are far away from the radar. These low-quality measurements significantly reduce the accuracy of metric localization. This module addresses this problem by taking the uncertainty into an optimization framework to find the best metric localization with minimum matching residuals.

Here we first convert the radar's raw measurements from polar coordinates to Cartesian coordinates for the ease of problem formulation. A point in radar submap can be expressed in the polar form as 
% \sout{$\hat{\bm{g}}_p = [\hat{r}, \hat{v}, \hat{\phi}]^\mathrm{T}$}
$\widetilde{\bm{g}}_p = [\widetilde{r}, \widetilde{v}, \widetilde{\phi}]^\mathrm{T}$, 
with the Gaussian noise $\bm{n}_p=[n_r,n_v,n_\phi]^\mathrm{T}$, \ie,
% \begin{equation}
% 	\hat{\bm{g}}_p = \bm{g}_p + \bm{n}_p,\ \bm{n}_p \in \mathcal{N}(0,\bm{\Sigma}_p), 
% 	\label{eq:6}
% \end{equation}

\begin{equation}
	\widetilde{\bm{g}}_p = \bm{g}_p + \bm{n}_p,\ \bm{n}_p \in \mathcal{N}(0,\bm{\Sigma}_p), 
	\label{eq:6}
\end{equation}
where $\bm{g}_p$ is the vector containing position, radial velocity, and azimuth angle. $\bm{\Sigma}_p=diag(\sigma_r^2, \sigma_v^2, \sigma_\phi^2)$ denotes the covariance of the measurements in the polar coordinate system. Note that we ignore the elevation angle accounting for its low resolution, but we still keep the z-axis by setting $z=0$ to make the radar outputs compatible with other 3D modalities. Thus the Doppler compensated radar target's nominal position $\widetilde{\bar{\bm{g}}}_c$ in Cartesian coordinate can be derived as:
\begin{equation}
	\begin{aligned}
		\widetilde{\bar{\bm{g}}}_c&=f(\bm{g}_p, \bm{n}_p) \approx f(\bm{g}_p) + \bm{J}\bm{n}_p, \\
		\bm{\Sigma}_c &= E[(\widetilde{\bar{\bm{g}}}_c - f(\bm{g}_p))(\widetilde{\bar{\bm{g}}}_c - f(\bm{g}_p))^\mathrm{T}] =\bm{J}\bm{\Sigma}_p \bm{J}^\mathrm{T}, 
	\end{aligned}
	\label{eq:7}
\end{equation}
where $f(\cdot)$ is the nonlinear function that transforms Doppler-compensated radar points from polar form to Cartesian form.
$\bm{J}$ denotes the Jacobian matrix \textit{w.r.t.} raw radar measurements,
\begin{equation}
	\begin{aligned}
		\bm{J}=\frac{\partial \widetilde{\bar{\bm{g}}}_c}{\partial [\widetilde{r},\widetilde{v},\widetilde{\phi}]^\top} = \begin{bmatrix}
			\cos{\widetilde{\phi}} &-\beta\cos{\widetilde{\phi}} &-(\widetilde{r}-\beta\widetilde{v})\sin\widetilde{\phi}\\
			\sin\widetilde{\phi} &-\beta \sin\widetilde{\phi} &(\widetilde{r}-\beta\widetilde{v})\cos\widetilde{\phi}\\
			0 & 0 & 0
		\end{bmatrix},
	\end{aligned} 
	\label{eq:8}
\end{equation}
where we set $\beta = \frac{f_c}{K}$ for simplicity.
\begin{figure}[t]
	\includegraphics[width=\linewidth]{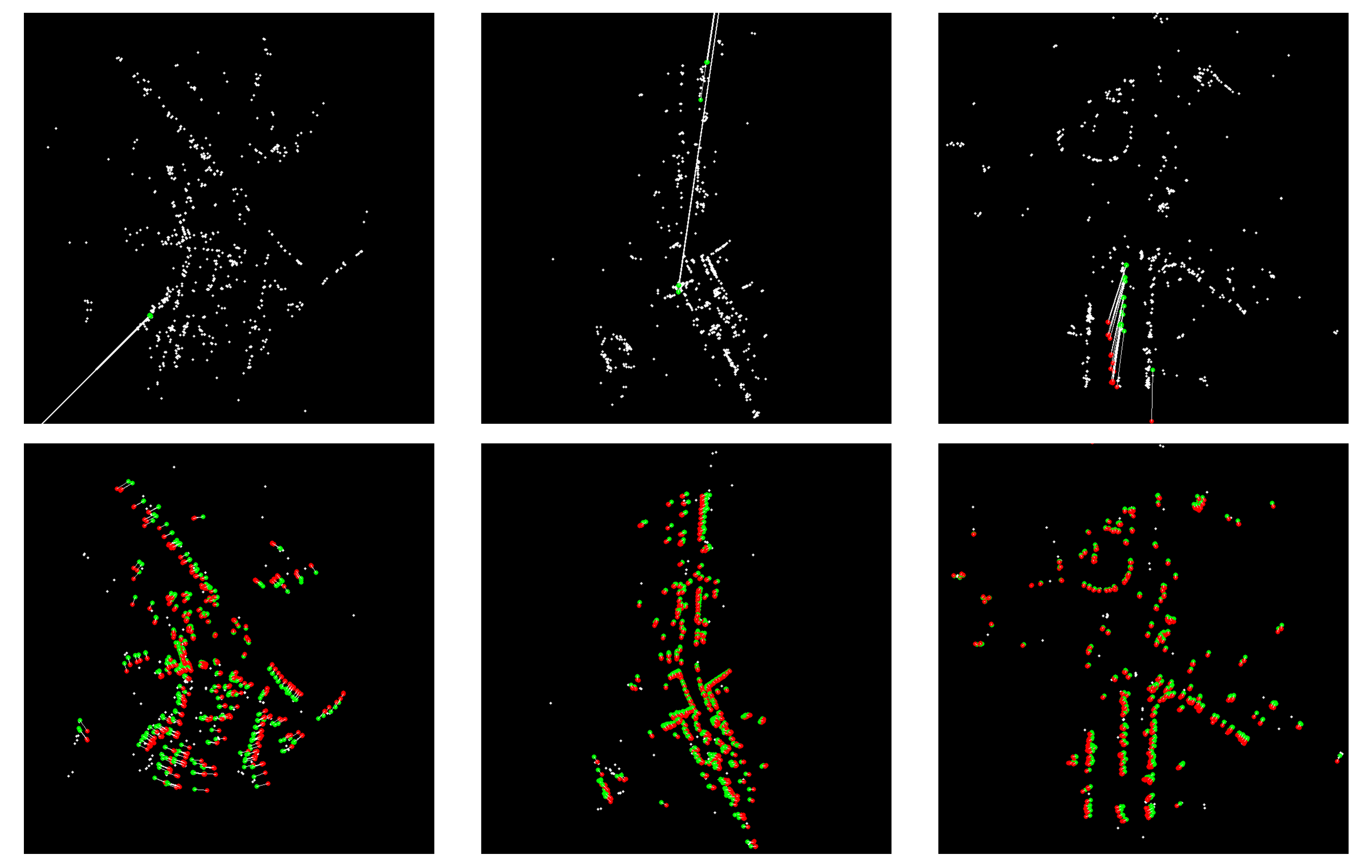}
	\caption{The matching results between two radar submaps using RSD \cite{cen2019radar}: The \textcolor[rgb]{0,0.75,0}{green} and \textcolor[rgb]{1,0,0}{red} dots represent the points in the current and previous radar submaps respectively. The Doppler-effect-free radar submaps (bottom row) can achieve more robust matching compared with those with Doppler distortion (top row).\vspace{-10pt}}
	\label{fig:matching}
\end{figure}

Fig.~\ref{fig:uncertainty} shows the uncertainty estimation of a point cloud. The gray areas around red points represent the uncertainty areas where the targets possibly to appear. As expected, those points further away from the sensor have larger uncertainties, and our system will punish these uncertain points by taking the Mahalanobis norm to be parts of the optimization objective.

%We should first know the point associations between two radar submaps.
In addition, good associations between radar submaps can significantly boost metric localization. We apply the state-of-the-art Radial Statistics Descriptor (RSD) \cite{cen2019radar} designed for the radar's point cloud to create a unique descriptor for each point in the submap. Then we take the \textit{K-Nearest Neighbor} (KNN) algorithm to find coarse associations based on the RSD. Furthermore, RANSAC~\cite{fischler1981random} excludes false matches from the initial associations. Fig.~\ref{fig:matching} shows that our Doppler-effect-free submaps can lead to more robust associations. The performance gain can be attributed to the Doppler compensation as the distortion destructs the geometric relationships between features, leading to the inconsistent RSD to the same point in different submaps. %, which will hamper correct association between radar point cloud.

%We denote the $i$\spth point in the current point set $\bm{Y}$ as $\hat{\bm{y}}_i\in\mathbb{R}^3$.
Given the associated point pairs, uncertainty estimation results, and the previous global positions of the radar, we can localize the current sensor's position by estimating the rigid body transformation between the current radar submap and the previous one, utilizing an optimization framework.  
We denote the $i$\spth point in the current point cloud $\bm{X}\in\mathbb{R}^3$ as $\widetilde{\bm{x}}_i$, and its corresponding point in the previous point cloud $\bm{Y}\in\mathbb{R}^3$ as $\widetilde{\bm{y}}_i$. For an arbitrary rigid transformation $\bm{T}_b^p\in SE(3)$, we define $d_i^{(\bm{T}_b^p)}=\widetilde{\bm{y}}_i-{\bm{T}_b^p}\widetilde{\bm{x}}_i$. The best transformation ${\bm{T}_b^p}^*$ between the current and the previous radar frames can be iteratively computed,  applying the Maximum Likelihood Estimate (MLE) scheme, \ie,
\begin{equation}
	{\bm{T}_b^p}^* = \arg\mathop{\max}\limits_{\bm{T}_b^p}\prod_{i=1}^{N}p(d_i^{(\bm{T}_b^p)})=\arg\mathop{\max}\limits_{\bm{T}_b^p}\sum_{i=1}^{N}log(p(d_i^{(\bm{T}_b^p)})),
\end{equation}
where $p(\cdot)$ is the distribution of $d_i^{(\bm{T}_b^p)}$. The above can be simplified to
\begin{equation}
	\begin{aligned}
		{\bm{T}_b^p}^* =\arg\mathop{\min}\limits_{\bm{T}_b^p}\frac{1}{2}\sum_{i=1}^{N}{d_i^{(\bm{T}_b^p)}}^\mathrm{T}(\bm{\Sigma}_{y,i}+\bm{T}_b^p\bm{\Sigma}_{x,i}{\bm{T}_b^p}^\mathrm{T})^{-1}d_i^{(\bm{T})}. 
	\end{aligned}
	\label{eq:10}
\end{equation}

We solve Eqn.~\eqref{eq:10} using the Levenberg-Marquardt (LM) optimization approach. A Cauchy robust loss function is also integrated to make the cost less sensitive to outliers \cite{barron2019general}. The transformation of the current body frame \textit{w.r.t.} world frame $\bm{T}_b^w$ can be expressed as: $\bm{T}_b^w = \bm{T}_p^w{\bm{T}_b^p}^*$, where $\bm{T}_p^w$ is the previous sensor's transformation to the global frame.

%% file: sections/experimental_setup.tex
\section{Evaluation}
\subsection{Experimental Setup}
Two datasets are used to evaluate the performance of \sysname, including the public nuScenes dataset \cite{caesar2020nuscenes}, and a synthetic dataset generated from the CARLA simulator \cite{dosovitskiy2017carla}. In Sec.~\ref{sub:Overall Performance on NuScenes}, we compare our approach with other radar-based metric localization methods. The ablation study is given in Sec.~\ref{sub:ablation study} to reveal the individual contribution of each component in \sysname. Finally, Sect.~\ref{sub:robust analysis} investigates the robustness of our system.
We now introduce the details of the two datasets as mentioned earlier.

\subsubsection{Real-world nuScenes Dataset}
The nuScenes is an autonomous driving dataset collected by a $32$-beam lidar, $5$ cameras, $5$ radars, and a GPS/IMU. The dataset has $242$ km driving data in total at an average speed of $16$ km/h. The radar used in this collection is the Continental ARS408-21 LRR automotive radar. This radar has a sampling rate of $13$ Hz, $250$ m sensing range, $0.39$ m range resolution, \ang{4.4} horizontal beamwidth, and \ang{3.2} azimuth resolution. These settings are representative of today's automotive radars operating in the frequency band $76 \sim 77$ GHz. To build up the metric localization dataset, we use the ground-truth position provided by nuScenes APIs and perform a KNN algorithm to obtain the nearest sample pairs across multiple driving rounds. In total, 21,746 pairs of radar submaps are found valid for our purpose of metric localization.

\subsubsection{Synthetic CARLA Dataset}
Most of the nuScenes dataset is collected at low ego-velocities, and the parameters of the onboard radar cannot be changed either. Therefore, it can hardly comprehensively demonstrate the robustness of our method in various settings.
To fully understand the robustness under a variety of ego-velocities and noise models, we synthesize more data by tuning radar parameters through the CARLA simulator.
Specifically, a car with a forward-facing radar is rendered in our experiment to collect data for localization. The maximum detection range and the radar's field of view (FOV) are set to be $100$ m and \ang{150}, respectively. %(c.f. Fig.~\ref{fig:CARLA render map})%. 
To examine different noise models of radar measurements, we generate three datasets with the noise following Gaussian distribution, Gamma distribution, and Student's t-distribution. Specifically, the noise to range measurements follows $n_{r_1}\sim\mathcal{N}(0,0.25^2)$, $n_{r_2} \sim \Gamma(0,0.25)$, and $n_{r_3}\sim 0.25\times t(100)$. The noise to angle measurements follows $n_{a_1}\sim\mathcal{N}(0,\ang{0.5}^2)$, $n_{a_2} \sim \Gamma(0,\ang{0.025})$, and $n_{a_3}\sim \ang{0.5}\times t(100)$. The noise to velocity measurements are $n_{v_1}\sim\mathcal{N}(0,0.1^2)$, $n_{v_2} \sim \Gamma(0,0.01)$, and $n_{v_3}\sim 0.1\times t(100)$. 
We simulate the Doppler effect with $\beta = 0.04$ (c.f. Eqn.~\eqref{eq:8}), referring to the technical parameters of ARS408-21.

% \begin{figure}[t]
% 	\includegraphics[width=\linewidth]{figure/Radar FOV.pdf}
% 	\caption{The radar (right) and the urban environment (left) that we render in the experiments.}
% 	\label{fig:CARLA render map}
% \end{figure} 

\begin{figure}[!t] \centering
	
	\includegraphics[width=0.93\columnwidth]{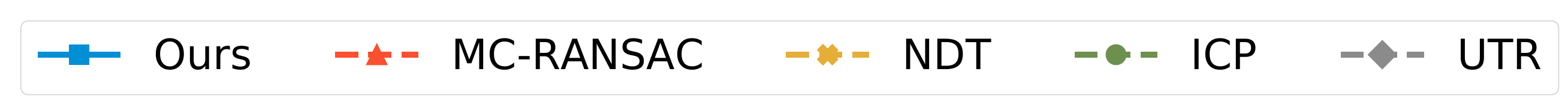}
	
	\subfigure[Translation error]{
		\label{fig:a}   
		\includegraphics[width=0.45\columnwidth]{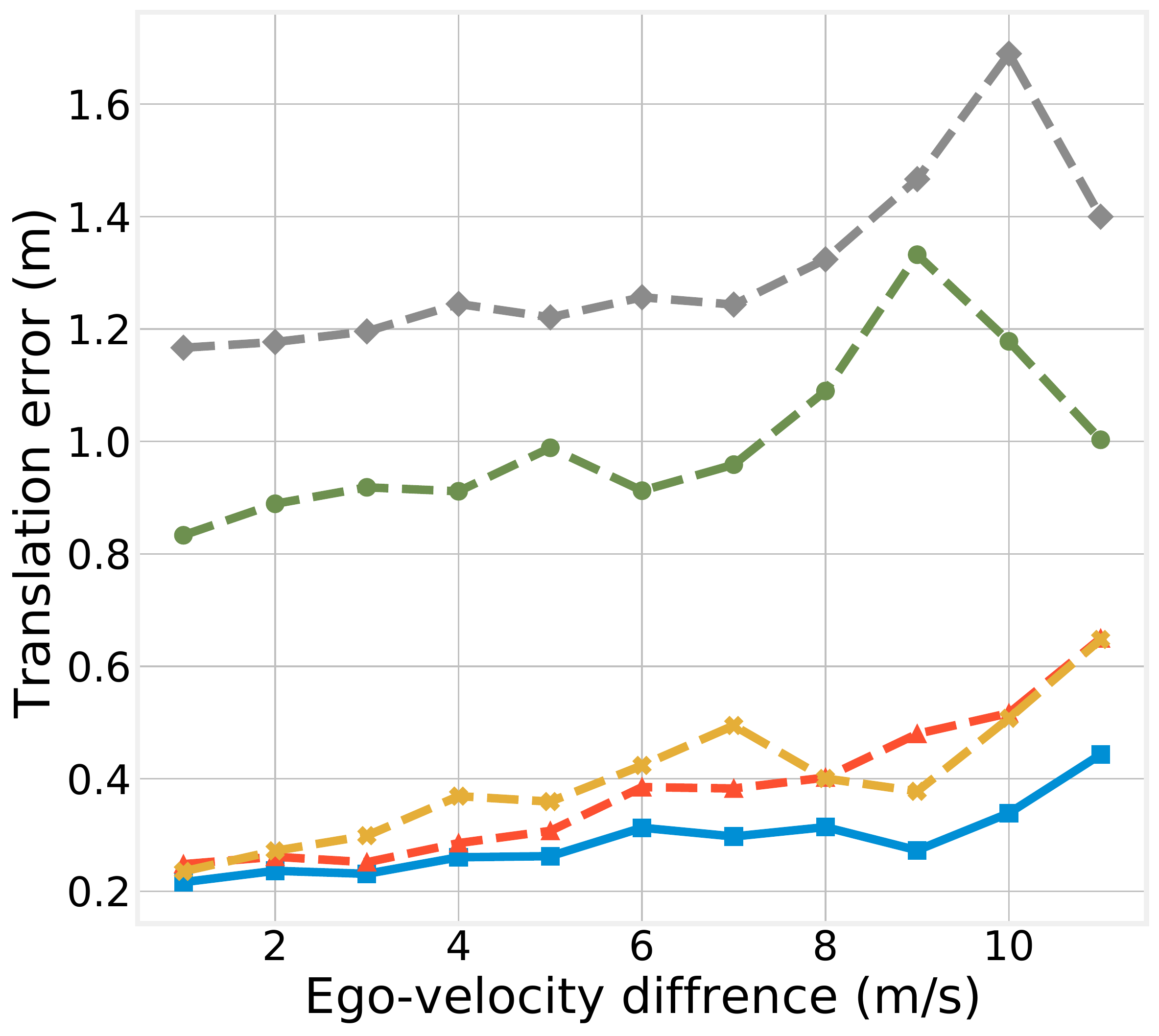}
	}
	\subfigure[Rotation error]{
		\label{fig:b}   
		\includegraphics[width=0.45\columnwidth]{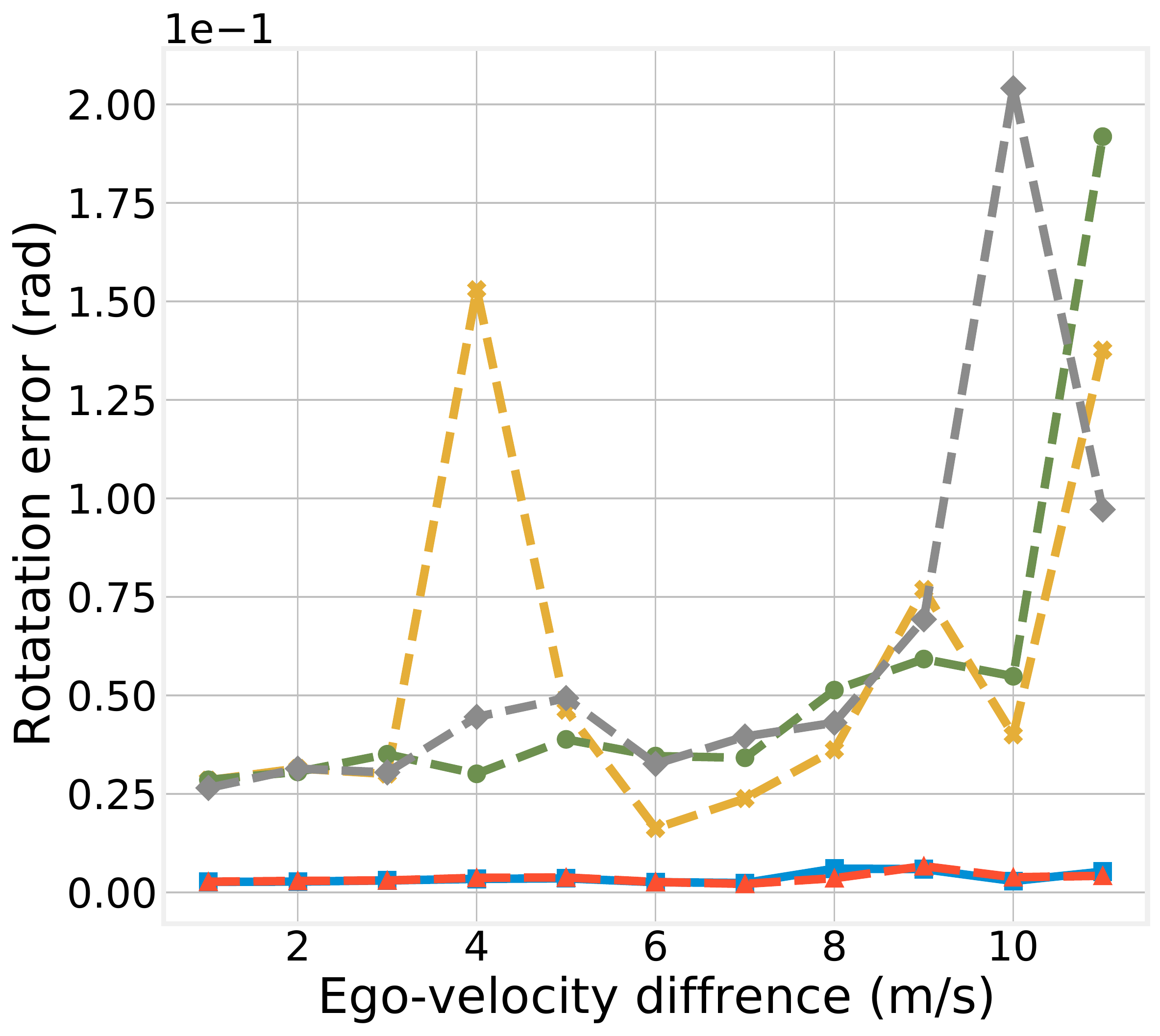}
	}
	\caption{Mean translation and rotation errors on the nuScenes dataset. \vspace{-25pt}}
	\label{fig:exp1}
\end{figure}

\begin{table}[!t]
	\centering
	\caption{Metric Localization Results on the nuScenes Dataset}
	\resizebox{\linewidth}{!}{%
        \begin{tabular}{|c|c|c|c|c|c|c|}
        \hline
        \multirow{2}{*}{Methods}                       & \multicolumn{3}{c|}{Translation Error (m)}       & \multicolumn{3}{c|}{Rotation Error (rad)}           \\ \cline{2-7} 
                                                   & Mean Err. & Max. Err. & Med. Err. & Mean Err. & Max. Err. & Med. Err. \\ \hline
        ICP\cite{besl1992method}  & 0.993    & 1.974    & 0.877     & 0.0536    & 0.3350   & 0.0204    \\ \hline
        NDT\cite{biber2003normal} & 0.400     & 0.957    & 0.330     & 0.0564    & 0.3857   & 0.0125    \\ \hline
        UTR\cite{barnes2020under} & 1.324     & 1.951    & 1.306    & 0.0576    & 0.1592   & 0.0276    \\ \hline
        MC-RANSAC\cite{burnett2021we} & 0.379          & 0.914          & 0.325          & 0.0018          & \textbf{0.0156} & 0.0012          \\ \hline
        Ours                                           & \textbf{0.290} & \textbf{0.873} & \textbf{0.243} & \textbf{0.0017} & 0.0159          & \textbf{0.0012} \\ \hline
        \end{tabular}%
	}
	\label{tab:1}
	\vspace{-10pt}
\end{table}

To create the submap pairs for metric localization, we conducted two rounds of radar data collection. The average speed of the former one is $72$ km/h, and the latter one is $40$ km/h, featuring a larger ego-velocity difference than that of nuScenes (around trifold larger). After applying the above KNN searching, 3,355 pairs of loop closure are eventually created for evaluation.

%(the version of metric localization with Doppler compensation)
\subsection{Overall Performance on nuScenes}
\label{sub:Overall Performance on NuScenes}
\subsubsection{Competing approaches}
We compare our approach with $4$ SOTA radar metric localization methods, including direct methods (convention ICP~\cite{besl1992method} and submap NDT~\cite{biber2003normal}), the learning-based method (UTR~\cite{barnes2020under}), and the feature-based method (MC-RANSAC~\cite{burnett2021we}). Note that the SOTA joint-Doppler-based NDT approach~\cite{kung2021normal} is designed for odometry rather than metric localization. Thus we implement a grid-based NDT~\cite{biber2003normal} for fair comparisons. Besides, since automotive radars are insusceptible to motion distortion, we only take the Doppler compensation component in MC-RANSAC for comparison. All methods use the radar submaps constructed by 10 consecutive radar scans. We set the same threshold of the translation error ($2.0$ m) to distinguish inliers and outliers for each method and only keep those inliers over all methods for performance analysis. 

%(2) lacking of good initial guess and sparse radar outputs can degrade the performance of direct-based and learning-based method respectively.
%can reduce estimation errors to some extent, it still suffers as the point cloud association in MC-RANSAC is based on the point clouds `polluted' by the Doppler effect.
\subsubsection{Results}
For ease of visualization, we present the results by grouping the velocity differences between two sensors and their corresponding translation and rotation errors on average. Fig.~\ref{fig:exp1} shows that the translation errors of ICP, NDT, UTR tend to grow larger as the velocity difference between two sensors increases. We attribute their poor performance to the lack of addressing Doppler distortion and sparse radar outputs. Note that MC-RANSAC's performance is very close to our approach, owning its attempts to handle the Doppler effects. However, its accuracy is still inferior because it cannot completely restore the Doppler distortion without velocity observations. Thus, the metric localization of MC-RANSAC takes the distorted submaps into the computation. As shown in Table~\ref{tab:1}, our approach can achieve the best translation results in all metrics by reducing $23.5\%$ mean error,  $4.5\%$ maximum error, and $25.2\%$ median error compared with MC-RANSAC.  In addition, our method can further provide $5.6\%$ improvements in the mean error of rotation estimation. The results demonstrate the effectiveness of our proposed explicit Doppler compensation component and uncertainty estimation process, which can significant boost the performance of radar metric localization. 
%\chris{Very bad description!!!! give a concrete number about exactly how much improvement you got. }

\subsection{Ablation Study}
\label{sub:ablation study}
We next study the individual contribution of each component in \sysname. Table~\ref{tab:2} shows the quantitative breakdowns.
It can be observed that the proposed Doppler compensation alone can significantly improve the translation and rotation estimation performance on average. The mean translation error reduces from $0.450$ m to $0.292$ m, and the mean rotation error also sees a $10.5\%$ decrease. However, the Doppler compensation does not curate the maximum error (\ie, the worst estimation case) because the maximum error is mostly caused by the outliers or extremely noisy measurements, which cannot be effectively mitigated by the compensation alone. 
In comparison, the proposed uncertainty estimation module can better deal with these outliers by giving more weights to those points with small uncertainties and consequently reducing the maximum translation error by $14.6\%$ and the maximum rotation error by $5.5\%$. Unfortunately, the uncertainty estimation cannot contribute to the mean or median errors as it is not bespoke designed to handle the dominant Doppler distortion in this experiment. 
In summary, thanks to the above complementary functionalities between the two components, it is not surprising that the best estimation results are achieved when using them in tandem.

\begin{table}[!t] \centering
	\caption{Ablation Study Results on the nuScenes Dataset}
	\scriptsize
	\begin{threeparttable}
		\resizebox{\linewidth}{!}{%
			\begin{tabular}{|c|c|c|c|c|c|c|c|}
				\hline
				\multicolumn{2}{|c|}{Components} & \multicolumn{3}{c|}{Translation Error (m)} & \multicolumn{3}{c|}{Rotation Error (rad)}  \\ \hline
				U.E.    & D.C.    & Mean Err. & Max. Err. & Med. Err.      & Mean Err. & Max. Err.        & Med. Err. \\ \hline
				-       & -       & 0.450     & 1.092    & 0.386          & 0.0019    & 0.0164          & 0.0012    \\ \hline
				$\surd$ & -       & 0.435     & 0.933    & 0.367          & 0.0017    & \textbf{0.0155} & 0.0012    \\ \hline
				-       & $\surd$ & 0.292     & 1.085    & \textbf{0.236} & 0.0018    & 0.0186          & 0.0012    \\ \hline
				$\surd$         & $\surd$        & \textbf{0.290}  & \textbf{0.873}  & 0.243  & \textbf{0.0017} & 0.0159 & \textbf{0.0012} \\ \hline
			\end{tabular}%
		}
		
		\begin{tablenotes}    
			\scriptsize         
			\item[1] U.E. - using uncertainty estimation.   
			\item[2] D.C. - using Doppler compensation.
		\end{tablenotes}          
	\end{threeparttable}
	\label{tab:2}
\end{table}

\subsection{Robust Analysis on CARLA Datasets}
\label{sub:robust analysis}
\begin{figure}[t]
	%    \center 
	\subfigure
	{\includegraphics[width=1.1in]{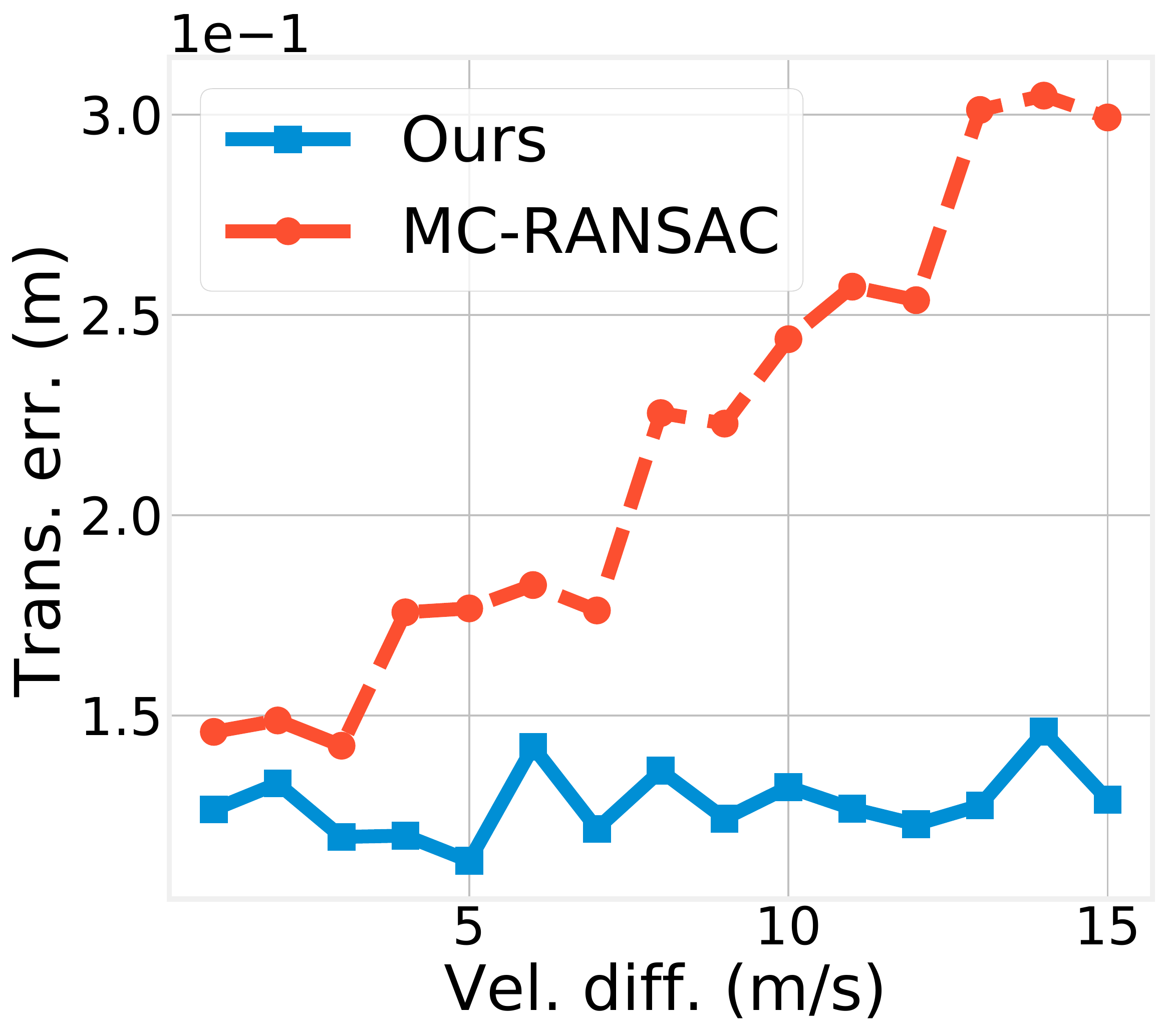}}
	\subfigure
	{\includegraphics[width=1.1in]{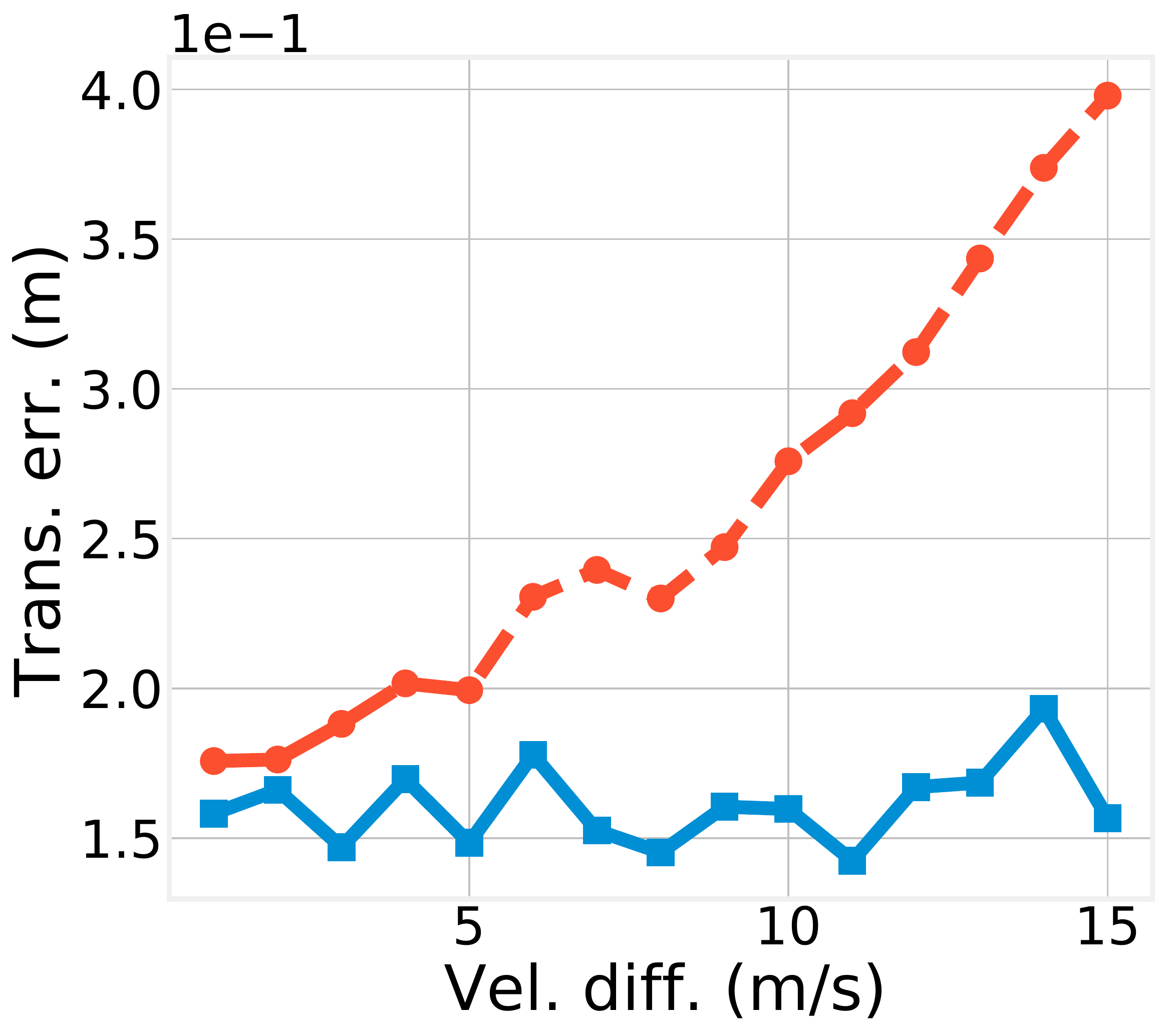}}
	\subfigure
	{\includegraphics[width=1.1in]{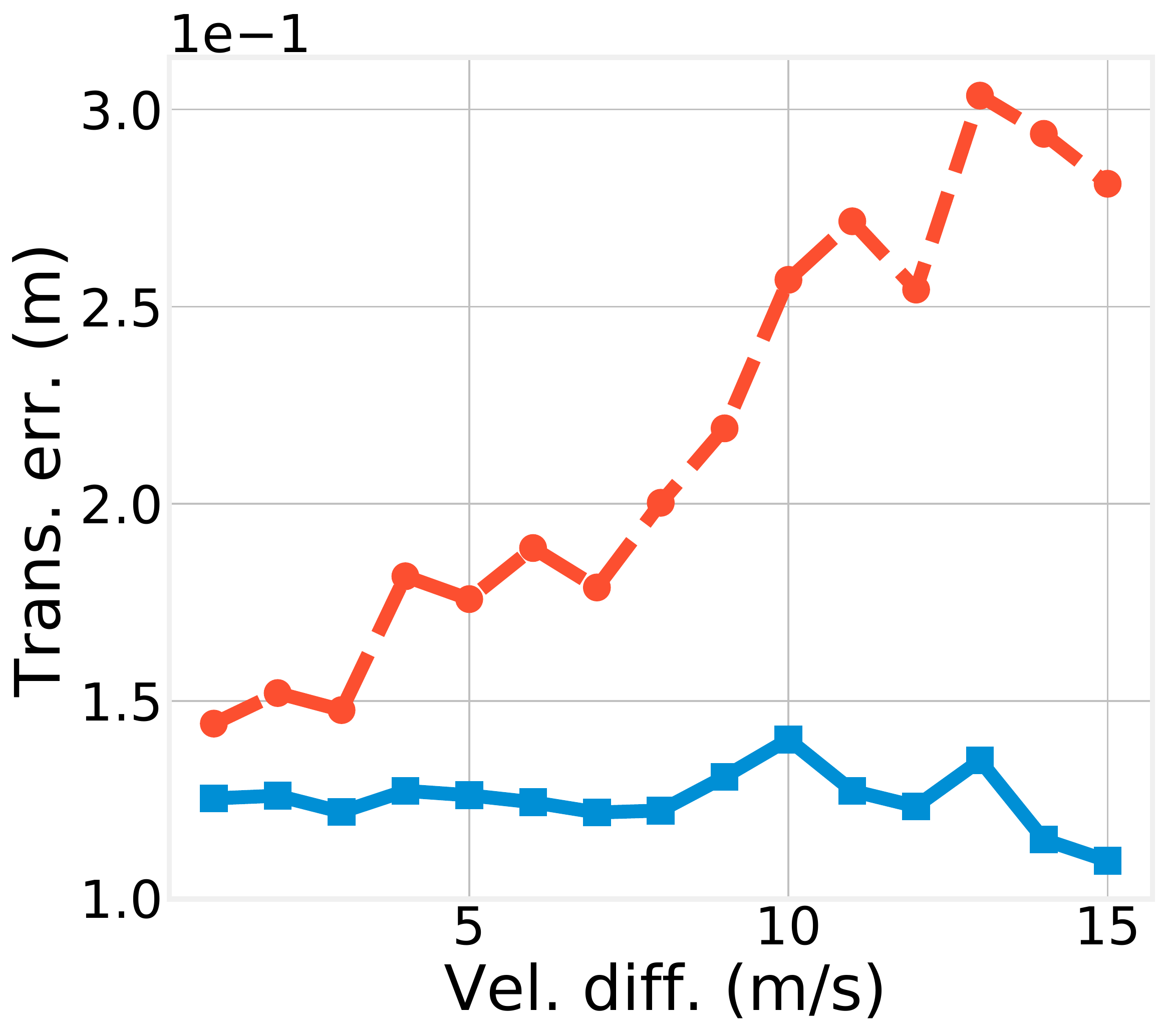}}
	\setcounter{subfigure}{0}
	\subfigure[Gaussian Noise]
	{\includegraphics[width=1.1in]{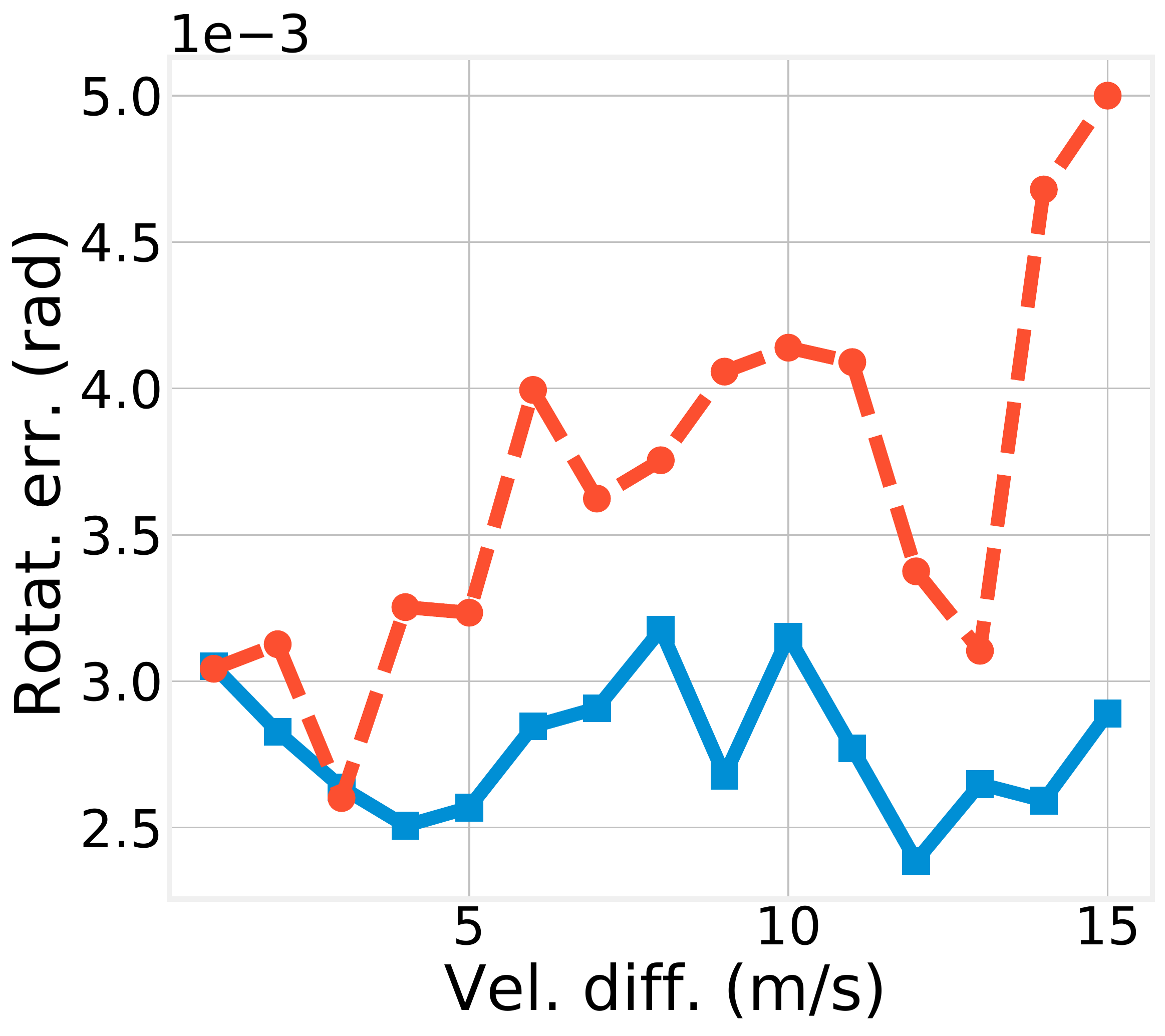}}
	\subfigure[Gamma Noise]
	{\includegraphics[width=1.1in]{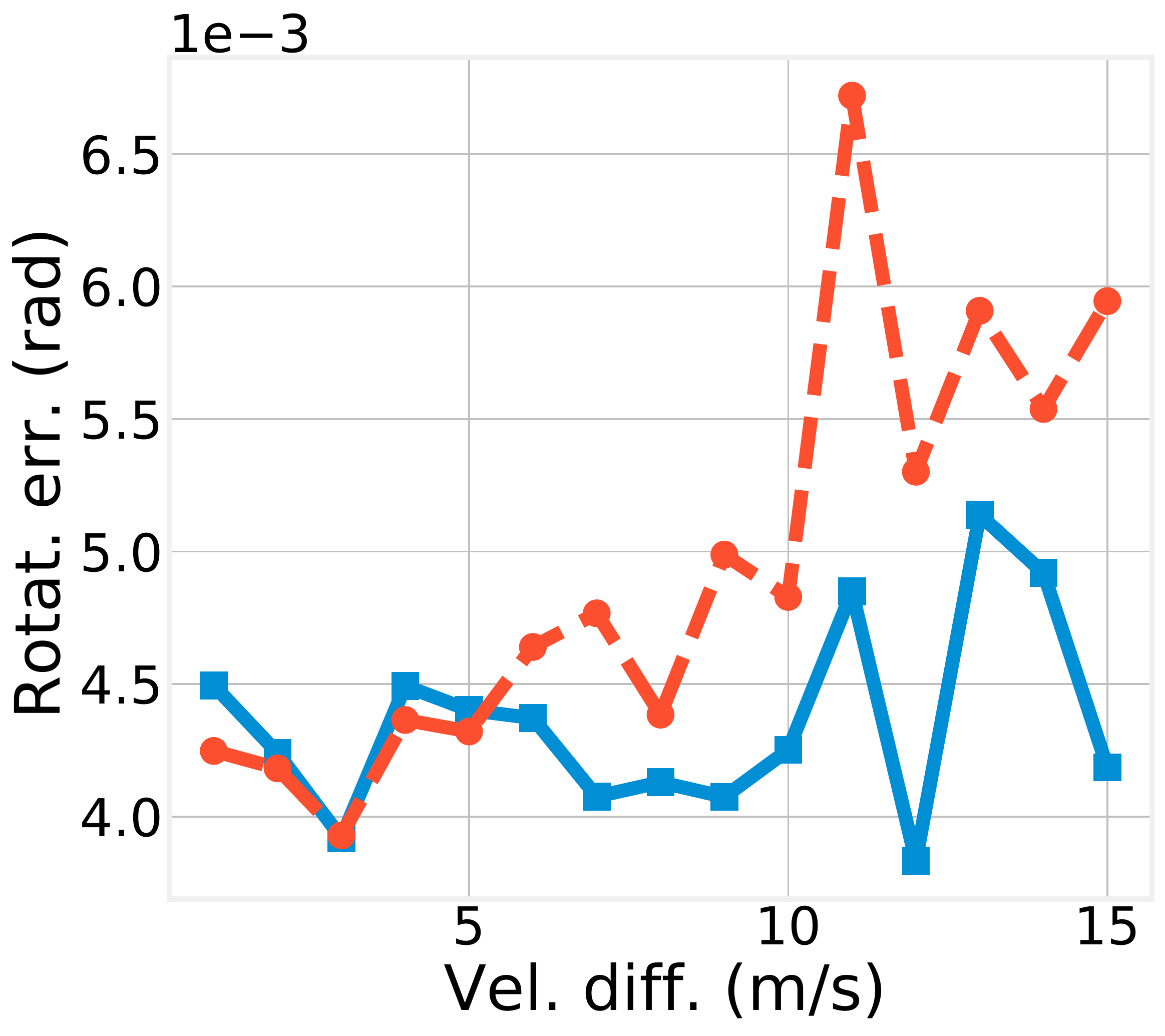}}
	\subfigure[Student's Noise]
	{\includegraphics[width=1.1in]{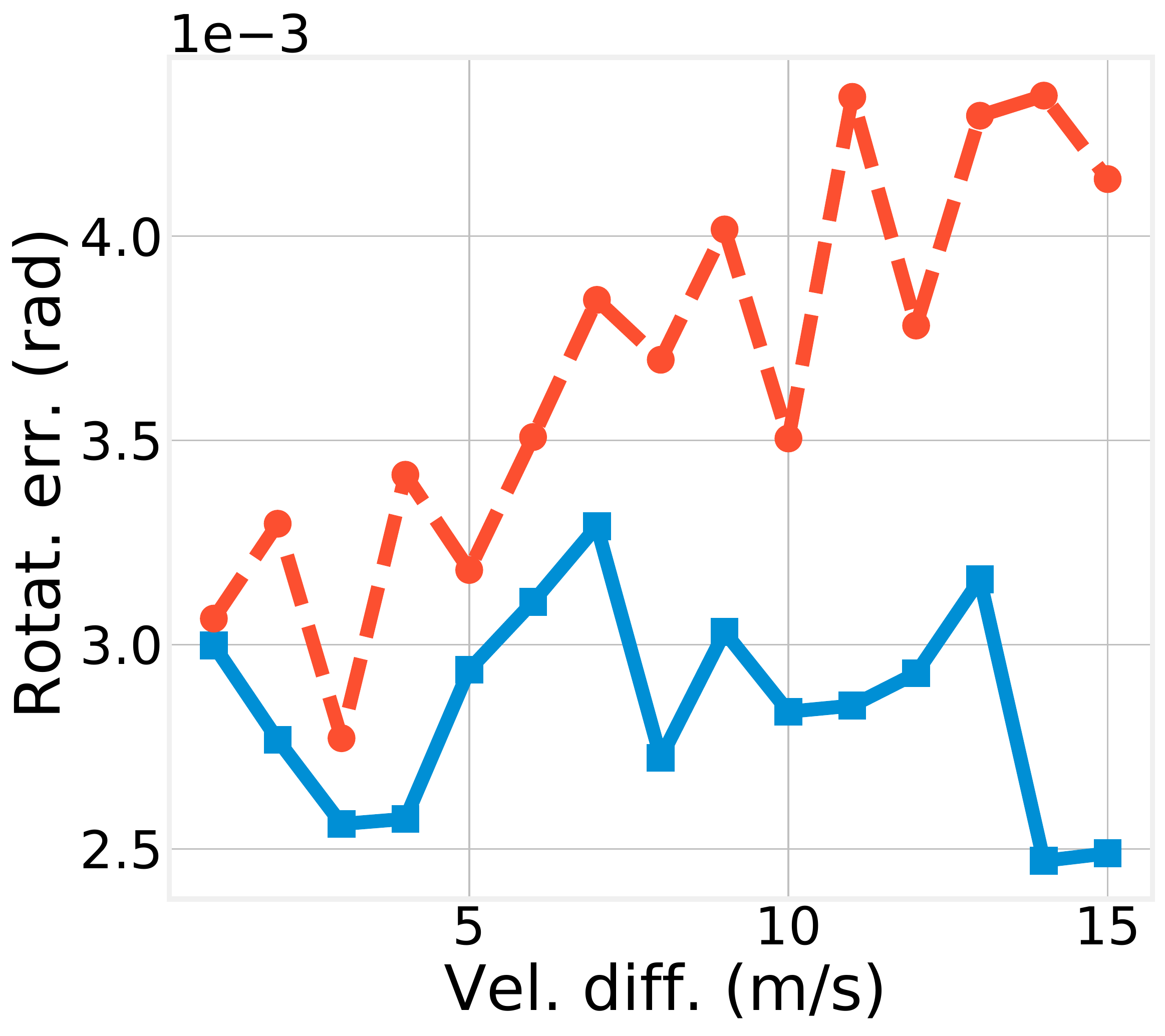}}
	\caption{The mean translation and rotation error on the synthetic dataset with Gaussian noise (left column), Gamma noise (middle column) and Student's noise (right column).}
	\label{fig:carla_fig}
\end{figure}

To investigate the robustness of \sysname under different types of noise, we further test our system, on the synthetic CARLA dataset. 
We compare our approach with MC-RANSAC on the aforementioned synthetic dataset and plot the mean translation and rotation errors in Fig.~\ref{fig:carla_fig}. As we can see, the translation and rotation errors of MC-RANSAC (the baseline second to the best on nuScenes) has a pronounced tendency to increase as the differences between ego-velocities become larger. In contrast, our method is robust to the Doppler effect and achieves a consistent performance under variant ego-velocities. Furthermore, our method has good stability under the effect of different types of noise. Specifically, our algorithm can outperform the MC-RANSAC by reducing $41.0\%$, $38.0\%$, $42.4\%$ translation errors and $25\%$, $12.2\%$, $22.2\%$ rotation errors with Gaussian noise, Gamma noise, and Student's noise, respectively.

In summary, The results demonstrate that \sysname ~has better robustness to ego-velocity differences and various types of noise.

\section{Conclusion}
This paper presents \sysname, an automotive mmWave radar-based metric localization framework, which addresses the Doppler distortion on radar scans. Our approach first proposes a novel method to explicitly compensate the Doppler distortion on radar ranging measurements to obtain more accurate submaps. Then it takes the radar measurement uncertainties into an optimization framework to further improve the metric localization. Extensive experiments using the real-world data from nuScenes and the synthetic data from CARLA demonstrate the effectiveness of our method in a variety scenarios. Our future work will integrate our method into an automotive radar-based SLAM system.